%% file: GEC.tex
\documentclass[11pt,a4paper]{article}
\usepackage[hyperref]{acl2019}
\usepackage{times}
\usepackage{latexsym}

\usepackage{url}
\usepackage{multirow}
\usepackage{graphicx}
\usepackage{amsmath}
\usepackage{verbatim}

\usepackage{color}
\usepackage{booktabs} %for nicer tables
\usepackage{subcaption}
\aclfinalcopy % Uncomment this line for the final submission
\definecolor{bblue}{HTML}{4F81BD}
\definecolor{rred}{HTML}{C0504D}

\title{Learning to combine Grammatical Error Corrections}

\author{ Yoav Kantor$^{*}$ \enspace Yoav Katz$^{*}$ \enspace Leshem Choshen\Thanks{Contributed equally} \enspace Edo Cohen-Karlik  \\ \enspace \textbf{Naftali Liberman} \enspace \textbf{Assaf Toledo} \enspace \textbf{Amir Menczel} \enspace \textbf{Noam Slonim} \\
\\IBM Research AI \\
   \texttt{\{yoavka,katz,leshem.choshen\}@il.ibm.com} \\
   \texttt{\{edo.cohen,naftali.liberman,assaf.toledo\}@ibm.com} \\ 
   \texttt{\{amir.menczel,noams\}}\texttt{@il.ibm.com}
  }

\date{}

\begin{document}
\maketitle
\begin{abstract}
  \input{abstract}

\end{abstract}

\input{intro}

\input{error_generation}
\input{systems}
\input{combine}

\input{experiments}
\input{discussion}

%\section*{Acknowledgments}

\bibliography{GEC}
\bibliographystyle{acl_natbib}

%\appendix

\end{document}

%% file: abstract.tex
The field of Grammatical Error Correction (GEC) has produced various systems to deal with focused phenomena or general text editing. We propose an automatic way to combine black-box systems. Our method automatically detects the strength of a system or the combination of several systems per error type, improving precision and recall while optimizing $F$ score directly.  We show consistent improvement over the best standalone system in all the configurations tested.
This approach also outperforms average ensembling of different RNN models with random initializations.

In addition, we analyze the use of BERT for GEC - reporting promising results on this end.  We also present a spellchecker created for this task which outperforms standard spellcheckers tested on the task of spellchecking.

This paper describes a system submission to Building Educational Applications 2019 Shared Task: Grammatical Error Correction\cite{bea2019}.

Combining the output of top BEA 2019 shared task systems using our approach, currently holds the highest reported score in the open phase of the BEA 2019 shared task, improving $F_{0.5}$ by 3.7 points over the best result reported.

%% file: intro.tex
\section{Introduction}\label{sec:intro}
Unlike other generation tasks (e.g. Machine Translation and Text Summarization), Grammatical Error Correction (GEC) contains separable outputs, edits that could be extracted from sentences, categorized \cite{Bryant2017AutomaticAA} and evaluated separately \cite{choshen2018automatic}. Throughout the years different approaches were considered, some  focused on specific error types \cite{rozovskaya2014correcting} and others adjusted systems from other tasks \cite{zhao2019improving}. While the first receive high precision, the latter often have high recall and differ in what they correct. To benefit from both worlds, pipelines \cite{rozovskaya2016grammatical} and rescoring hybrids \cite{grundkiewicz2018near} were introduced. Another suggested method for combining is average ensembling \cite{JunczysDowmunt2018ApproachingNG}, used when several end to end neural networks are trained.

As single systems tend to have low recall \cite{choshen2018inherent}, pipelining systems may propagate errors and may not benefit from more than one system per error. Rescoring reduces recall and may not be useful with many systems \cite{grundkiewicz2018near}.  We propose a new method for combining systems (\S \ref{sec:comb}) that can combine many systems and relies solely on their output, i.e., it uses systems as a black-box. We show our system outperforms average ensembling, has benefits even when combining a single system with itself, and produces the new state of the art by combining several existing systems (\S \ref{sec:experiments}).

To develop a system we trained GEC systems and gathered outputs from black-box systems (\S \ref{sec:systems}). One of the most frequent error types is spelling errors, we compared off of the shelf spellcheckers, systems developed for this error type specifically, to a new spellchecker (\S \ref{subsec:spell}), finding that our spellchecker outperforms common spellcheckers on the task of spellchecking.

Another system tested was modifications of BERT \cite{devlin2018bert} to correct errors, allowing for less reliance on parallel data and more generalizability across domains (\S \ref{subsec:bert}). 

Lastly, we tested generating synthetic errors \cite{felice2014generating} as a way to replace data in an unsupervised scenario. While finding that mimicking the error distribution and generating errors on the same domain is better, we did not eventually participate in the low-resource track.

%% file: error_generation.tex
\section{Data}
\subsection{Preprocessing}
Many systems assume the input is standard untokenized English sentences. In these cases, we detokenized the input data sets and then tokenized again to perform the combination and evaluation steps.  For training the Nematus network, we passed the data tokenization and truecasing \cite{koehn2007moses} and trained BPE \cite{sennrich2015neural}.
\subsection{Synthetic Error Generation} \label{sec:synth}
Generating training data for the GEC problem is expensive and slow when done manually by human annotators. Most machine-learning based systems today benefit from the quantity and richness of the training data, therefore, generating synthetic data has a lot of potential, as was also shown in previous work \cite{felice2014generating}.
We generate data with errors by applying corrections backwards. Meaning, if a correction adds a missing word $X$ to a sentence, to produce the corresponding error we remove $X$ from a sentence. And if a correction removes a redundant word $X$ from a sentence, to produce the corresponding error we add word $X$ in a random location in a sentence. And if a correction replaces word $X$ with word $Y$ in a sentence, to produce the corresponding error we replace word $Y$ with word $X$ in a sentence.
In order to preserve the distribution of errors as found in the W\&I+LOCNESS train data set, we analyze it and measure the distribution of corrections in it. We measure the distribution of number of corrections in a sentence and distribution of specific corrections. Using these distributions and a corpus of gold (correct) sentences we produce errors with similar distributions.
We first randomly select the number of corrections in a sentence according to the distribution measured before. Then, we randomly select specific corrections according to the distribution of corrections. We then find all sentences where all corrections can be applied backwards and pick one of them randomly. Lastly, we generate the errors in the sentence and add the gold sentence and error sentence to corresponding output files.

%% file: systems.tex
\section{Systems} \label{sec:systems}
\subsection{Constructing a spellchecker}\label{subsec:spell}
Many tools are available for spelling correction. Yet, with a few heuristics we managed to get a comparatively high result. As by Errant \cite{Bryant2017AutomaticAA}, our spellchecker receives a better $F_{0.5}$ score of spelling (type R:SPELL) than other leading open-source spell-checkers.  A comparison can be found at \S \ref{subsec:spell_exp}. 

Our method of correcting spelling mistakes is as follows. As a preprocessing stage, we go over a large monolingual corpus - specifically a 6 million sentences corpus taken from books in project Gutenberg\footnote{\url{https://www.gutenberg.org}}. We count the number of occurrences of each word (in it's surface form), skipping words with less than 3 characters and words that are not composed exclusively of letters.
We also use an English dictionary (both US and GB) from LibreOffice site \footnote{\url{https://cgit.freedesktop.org/libreoffice/dictionaries/tree/en}} for enriching our data with English words that are not in our books corpus.
When correcting a sentence, we find words that are not in our word-count (or in it and have a count below 3) nor in the Dictionary. Skipping words with digits or if it was all upper case. These words are suspected to be misspelled and we try to correct them.

For every misspelled word we try to find a replacement word by going over the words in the word-count data (words with count greater than 20) in a descending order of occurrences. For each suggested word, we check if it can be considered as a correction for the misspelled word by two methods. First, we check if the original word and the candidate correction differ from each other by swapping two characters. If not, we calculate the distance between the two words using Levenshtein distance \cite{Levenshtein1966BinaryCC} and check if the distance is 1. We return the most frequent word that satisfies one of these conditions . If no candidate is found, we do the same with all words in the dictionary in a lexicographical order. If still no candidate is found, we check if we can split the misspelled word into two words that are in our word-count data or in the dictionary.

\subsection{Nematus}
We trained 4 neural machine translation systems based on Nematus \cite{sennrich2017EACLDemo} Transformer \cite{Vaswani2017Attention} implementation. All parameters used are the ones suggested for the 2017 Workshop on Machine Translation \footnote{\url{https://github.com/EdinburghNLP/wmt17-transformer-scripts}}. As training data we used all the restricted data, i.e., FCE \cite{dale2011helping}, LANG8 \cite{mizumoto2011mining}, NUCLE \cite{dahlmeier2013building} and W\&I+LOCNESS \cite{bea2019, Granger1998TheCL} (upsampled 10 times). Each of the four trained models was regarded as a separate correction method and all systems were combined using our method (\S \ref{sec:comb}), this was especially beneficial as ensembling is not yet implemented for the transformer. See \S \ref{subsec:ensemble_compare} for comparison of the two ensembling methods over RNN based Nematus.

\subsection{Off the shelf}
\paragraph{LanguageTool.}
\href{https://github.com/languagetool-org/languagetool}{LanguageTool} is a free grammar correction tool mainly based on spellchecking and rules.  We used language tool programmatic API to obtain all the possible corrections and applied all the suggestions.
\paragraph{Grammarly.} 
\href{https://www.grammarly.com/}{Grammarly} is the company owning the world leading grammar correction product, as such it is the obvious candidate to be used as a component and to assess the potential of combining black box systems. We used their free web interface to correct the dev and test sets.  Grammarly does not support a programmatic API, so this process was manual.  We uploaded the texts after detokenization into the web interface.   For each suggested correction, we took the top prediction without human discretion. The reason to choose the top prediction was to allow combining using a single reference of Grammarly.
\paragraph{Spelling correction.}
We tested \href{https://www.abisource.com/projects/enchant/}{Enchant},  \href{https://github.com/bakwc/JamSpell}{JamSpell} and \href{https://github.com/barrust/pyspellchecker}{Norvig} spellcheckers, finding our spellchecker outperforms those in terms of spelling correction (See \S \ref{sec:experiments}).
\subsection{BERT} \label{subsec:bert}
BERT (Bidirectional Encoder Representations from Transformers) \cite{devlin2018bert} is a language representation model. BERT is extremely effective in general purpose tasks, among its virtues, BERT holds a syntactic understanding of a language \cite{DBLP:journals/corr/abs-1901-05287}. Initial pre-training of BERT was performed over a large corpora jointly on two tasks: (1) \textit{Masked Language Model} - randomly replace words with a predefined token, \verb|[MASK]|, and predict the missing word. (2) \textit{Next Sentence Prediction} - given a pair of sentences \verb|A| and \verb|B|, does sentence \verb|B| follow sentence \verb|A|.

Our general approach for using BERT to solve the GEC task is by iteratively querying BERT as a black box language model, reminding former use of language models \cite{dahlmeier2012beam,bryant-briscoe-2018-language}.
To detect missing words we add \verb|[MASK]| between every two words, if BERT suggests a word with high confidence, we conclude that this word is missing in this gap. To detect unnecessary words, we replace words with the \verb|[MASK]| token and if all the suggestions returned from BERT have a low probability, we conclude that the masked word was unnecessary. For replacing words, we perform the same procedure by replacing each word with \verb|[MASK]| and checking if BERT returns a different word with high probability.

The described process produces many undesired replacements/deletions due to BERT's versatile nature, for example, given a sentence such as:
\begin{center}
    \verb|There are few ways to get there.|
\end{center}
BERT may suggest replacing \verb|few| with \verb|many|. Such a replacement preserves the grammatically soundness of the sentence, but alters the semantic meaning. Hence, although possibly improving fluency, arguably the true goal of GEC \cite{napoles2017jfleg}, this behaviour does not align with the goals of GEC requiring semantic preservation \cite{choshen2018reference}. In order to focus the exploration space of BERT's suggestions, we limit replacements/deletions to operate within a predefined \textit{word set}. The word sets considered included syntactically interchangeable words, often sharing some semantic properties. When considering a removal correction, we remove a word only if the returned values from BERT are not in the same word-set as the replaced word. Replacement is allowed only within the same word set.
For example, a typical mistake which occurred frequently in the dataset is wrong usage of determiners such as \textbf{a} and \textbf{an}, given the word set $\lbrace a,an \rbrace$ and the sentence:
\begin{center}
    \verb|Is that a armadillo?|
\end{center}
The mechanism described limits the replacement correction options to suggest making a replacement-correction of \verb|a| with \verb|an| to result with the corrected sentence
\begin{center}
    \verb|Is that an armadillo?|
\end{center}

At each iteration of this process, a correction (addition/replacement/deletion) is performed and the resulting sentence is then used as the input to the next iteration.
Each replacement/addition of the \verb|[MASK]| token is a single candidate for a specific correction. Given an input sequence, each possible correction gives rise to a different candidate which is then sent to BERT. The most probable correction (above a minimal threshold) is then selected, this process accounts for one iteration. The resulting sentence is then processed again and the best correction is chosen until all corrections have a low probability in which case the sentence is assumed to be correct.

The above mechanism with threshold values between 0.6 and 0.98 did not yield satisfying results. For this reason, in the submitted system we limit the mechanism significantly, ignoring additions and deletions to focus solely on the replace corrections. Word sets were chosen from the most frequent errors in the training data across different error types (excluding punctuation marks R:PUNCT).

Another approach for using BERT is by fine-tuning BERT to the specific data at hand. Since the GEC task is naturally limited to specific types of errors, we fine-tuned the \textit{Masked Language Model} task using synthetic data. Instead of randomly replacing words with the \verb|[MASK]| token, we replace only specific words in a distribution which mimics the training data. This process should create a bias in the language model towards the prediction of words which we want to correct. Unfortunately, these efforts did not bear fruit. The authors believe a more extensive exploration of experimental settings may prove beneficial.

%% file: combine.tex
 \section{Combining systems} \label{sec:comb}
 Combining the output of multiple systems has the potential to improve both recall and precision. Recall is increased because typically different systems focus on different aspects of the problem and can return corrections which are not identified by other systems \cite{Bryant2017AutomaticAA}. Precision can be increased by utilizing the fact that if multiple systems predict the same annotations, we can be more confident that this correction is correct.  
 
 The outputs of Seq2Seq models, differing in training parameters, can be merged using an ensemble approach, where the predictions of the models for each possible word in the sequence are used to compute a merged prediction.  It was shown that even an ensemble of models trained with the same hyperparameters but with different instances of random initialization can yield benefit \cite{JunczysDowmunt2018ApproachingNG}.
 
 The idea of automatically combining multiple system outputs is not new to other fields and was successfully used in the Named Entity Recognition (NER) and Entity linking (EL) tasks.
 \citet{jiang2016evaluating} evaluated multiple NER systems and based on these results, manually selected a rule for combining the two best systems, building a hybrid system that outperformed the standalone systems.
 \citet{ruiz2015combining} used the precision calculated on a training corpus to calculate a weighted vote for each EL output on unseen data.
 \citet{dlugolinsky2013combining} used decision tree classifier to identify which output to accept. They used a feature set based on the overall text, NE surface form, the NE type and the overlap between different outputs. In GEC, combining was also proposed but was ad-hoc rather than automatic and general. Combining was done by either piping \cite{rozovskaya2016grammatical}, where each system receives the output of the last system, or correction of specific phenomena per system \cite{rozovskaya2011Algorithm}, or more involved methods tailored to the systems used \cite{grundkiewicz2018near}. This required manual adjustments and refinements for every set of systems.

 Evaluating by a corpus level measure such as $F$ score renders combining systems difficult. Systems developed towards $F_{0.5}$ tend to reduce recall improving precision \cite{choshen2018inherent}, while avoiding catastrophic errors \cite{choshen2018reference} this behaviour might reduce the flexibility of the combination. It is possible to tune systems to other goals (e.g. recall) \cite{grundkiewicz2018near} and thus achieve more versatile systems, but that is not the case when using black-box systems, and hence left for future inspection.
 %On the other hand, combining in a way that optimizes $F_{0.5}$ is a challenge.  While it is possible to change the weights or loss function to bias toward  created precision or greater recall, the optimizing the overall Fbeta measure requires a global view of all annotation types, their recall and precision results.   This is because Fbeta is not a linear measure, the  gradient of f-beta depends on the global recall and precision values.
 
 \paragraph{System pair.} We propose a method to combine multiple systems by directly optimizing $F_{\beta}$ for a chosen $\beta$, in the field 0.5 is usually used. We begin by considering a combination of two systems
 
 \begin{enumerate}
 % If we will need to shorten the paper, we can shorten:
 \item Given a development set, where $E$ are the sentences with errors and $G$ are the gold annotations, generate $M^2_{gold}$ file, which contains all the gold corrections to the sentences.
 \item Correct $E$ with each of the systems, to receive corrected sentences hypothesis $H_i$.
 \item Generate $M^2_{i}$ for each system $i$ by comparing the systems' output $H_{i}$ and the $E$ input. 
 \item Split the annotations of the systems into three subsets: $H_{1 \setminus 2}$ - all the suggested annotations of $system1$ which were not suggested by $system2$; $H_{2 \setminus 1}$ - all the suggested annotations of $system2$ which were not suggested by $system1$; and $H_{1 \cap 2}$ - all the suggested annotations in common.
 \item Generate $M^2$ files for each of the three sets: $M^2_{1 \setminus 2}$, $M^2_{1 \setminus 2}$, $M^2_{1 \cap 2}$.
 \item Evaluate the performance on each of the three subsets of annotations, split by error type, by comparing $M^2_{subset}$ with $M^2_{gold}$. For each subset and each error type, we obtain $TP^{error-type}_{subset}$,  $FP^{error-type}_{subset}$ ,  $FN^{error-type}_{subset}$.
 \item Define selection variables $S^{error-type}_{subset}$ which determine the probability an edit of the specific error type in a specific subset of edits will be used. According to the way subsets were built, each edit corresponds to exactly one subset (e.g. $1 \setminus 2$).  
 \item For all error types and subset of edits, compute the optimal selection variables $S^{error-type}_{subset}$ that maximize $f_{\beta}$ by solving
 \begin{equation*}
 \begin{split}
     0 &\leq S^{error-type}_{subset} \leq 1\\
     total &= \sum_{t \in error-type}  TP^{t}_{1 \cap 2} + FN^{t}_{1 \cap 2}\\
     TP &= \sum_{t \in error-type,s \in subset} TP^{t}_{s} * S^{t}_{s}\\
     FP &= \sum_{t \in error-type,s \in subset} FP^{t}_{s} * S^{t}_{s}\\
     FN &= total - TP\\
     Sopt &= \arg \max_{S} f_{\beta}(TP, FP, FN)
 \end{split}
 \end{equation*}
 This is a convex optimization problem with linear constraints and pose no difficulty to standard solvers.  
 \end{enumerate}
 
 $Sopt^{error-type}_{subset}$ need not be integer, although in practice they usually are.  \footnote{Non integer value can occur when a $0$ value yields high precision and low recall, and a $1$ value yields low precision and high recall. In this case, randomly selecting a subset of the corrections will yield a medium recall and medium precision, which maximizes $f_{\beta}$}.  In our submission, for simplicity, we avoid these cases and round $Sopt^{error-type}_{subset}$ to nearest integer value (either 0 or 1).  But our implementation allows sampling.
 
 A major concern is to what extent does the precision and recall statistics per error type and subset on the development set represent the actual distribution expected during inference on unseen data.  Assuming the development set and the unseen are sampled from the same distributions, the confidence is correlated with the number of samples seen for each error-type and subset.   

% if we need space, this could be deleted, generalization error
Assuming errors come from a binomial distribution, we try to estimate the conditional probability 
 $P\left( \left| prec_{test} - prec_{dev}\right| < 0.15 \mid prec_{dev} \right)$.
 Given more than 20 samples, the probability for 15\% difference in development and test precision is 14.5\%, and if there are 50 samples, this probability drops to 2.8\%.
 In the experiments, we ignore error-types where there are less than 2 samples.

 The process of correcting an unseen set of sentences $T$ is as follows:
\begin{enumerate}
\item Correct $T$ by every system $i$, to receive corrected sentences hypothesis $H_{i}$ .
 \item Generate $M^2_{i}$ files for each system by comparing the systems' output $H_{i}$ and the $T$ input. 
 \item Split the annotations of the systems into three sets:  $H_{1 \setminus 2}$ , $H_{2 \setminus 1}$ , and $H_{1 \cap 2}$ .
 \item Generate $M^2$ files for each of the three sets: $M^2_{1 \setminus 2}$, $M^2_{2 \setminus 1}$, $M^2_{1 \cap 2}$.
 \item Remove all annotations from the $M^2$ files for which  $Sopt^{error-type}_{subset} = 0$.
 \item Merge all the annotations from the modified $M^2_{1 \setminus 2}$, $M^2_{2 \setminus 1}$, and $M^2_{1 \cap 2}$ files to create $M^2_{final}$. If there are overlapping annotations - we currently select an arbitrary annotation.
 \item Apply all the corrections in $M^2_{final}$ to $T$ and receive the final output.
\end{enumerate}

In Table \ref{ta:example_selection}, we present the results of the most frequent error types when combining two systems, Nematus and Grammarly.
As expected, the precision on corrections found by both systems is significantly higher than those found by a single system.   For correction type 'R:OTHER', for example, the precision on common corrections is 0.67, compared to 0.17 and 0.28 of the respective standalone systems. Therefore, the optimal solution uses only the corrections produced by both systems. 
We can also see that in some error types (e.g., R:SPELL or R:DET) the precision of  corrections identified by the Nematus system is low enough that the optimization algorithm selected only the corrections by Grammarly.

\begin{table*}[htb]
\resizebox{\textwidth}{!}{%
\begin{tabular}{ |l|l||c|c|c||c|c|c||c|c|c|}
\hline
error-type & Frequency & 
$S_{1 \setminus 2}$ & $P_{1 \setminus 2}$ & $R_{1 \setminus 2}$ & 
$S_{1 \cap 2}$ & $P_{1 \cap 2}$ & $R_{1 \cap 2}$ & 
$S_{2 \setminus 1}$ & $P_{2 \setminus 1}$ & $R_{2 \setminus 1}$ \\
\hline
R:PUNCT&\%4&1.0&0.47&0.15&0.0&0.0&0.0&1.0&0.4&0.01 \\
U:DET&\%4&1.0&0.38&0.07&1.0&0.77&0.15&1.0&0.51&0.2 \\
R:VERB&\%5&1.0&0.5&0.02&0.0&1.0&0.01&1.0&0.57&0.02 \\
M:DET&\%5&0.0&0.29&0.05&1.0&0.68&0.12&1.0&0.4&0.31 \\
R:ORTH&\%5&0.0&0.28&0.22&1.0&0.86&0.13&1.0&0.46&0.18 \\ 
R:SPELL&\%5&0.0&0.32&0.04&1.0&1.0&0.11&1.0&0.66&0.65 \\ 
R:VERB:TENSE&\%5&1.0&0.54&0.15&0.0&0.0&0.0&0.0&0.0&0.0 \\ 
R:PREP&\%6&1.0&0.37&0.07&1.0&0.72&0.07&1.0&0.56&0.1 \\
R:OTHER&\%11&0.0&0.17&0.02&1.0&0.67&0.02&0.0&0.28&0.04 \\
M:PUNCT&\%15&1.0&0.55&0.17&1.0&0.68&0.06&1.0&0.38&0.12 \\ 
\hline
\end{tabular}
}
\caption{Combination statistics of the most common error types over two systems - Nematus and Grammarly}  \label{ta:example_selection}

\end{table*}

 \paragraph{Multiple systems.} When $N>2$ systems are available, it is possible to extend the above approach by creating more disjoint subsets, which include any of the $2^N$ subsets of corrections. When $N$ is large, many of these subsets will be very small, and therefore may not contain meaningful statistics.  We propose an iterative approach, where at each step two systems are combined.  The results of this combination can be then combined with other systems. This approach works better when the development set is small, but can also suffers from over-fitting to the dev set, because subsequent combination steps are performed on the results of the previous merges steps, which were already optimized on the same data set.

%% file: experiments.tex
\section{Experiments} \label{sec:experiments}
As our system is based on various parts and mainly focuses on the ability to smartly combine those, we experiment with how each of the parts work separately. A special focus is given to combining strong components, black-box components and single components as  combining is a crucial part of the innovation in this system.
\subsection{Spell checkers' comparison} 
\label{subsec:spell_exp}
We've compared our home-brewed spell-checker with JamSpell\footnote{\url{https://github.com/bakwc/JamSpell}}, Norvig\footnote{\url{https://github.com/barrust/pyspellchecker}} and ENCHANT\footnote{\url{https://github.com/AbiWord/enchant}}. When comparing the results over all error categories, our spell-checker has relatively low results (See Table \ref{ta:sp_gec}). However, when comparing the results in spelling (R:SPELL) category alone, our spell-checker excels (See Table \ref{ta:sp_sp}).

\begin{table}[htb]
\resizebox{\columnwidth}{!}{%
\begin{tabular}{|l|l|l|l|l}
\cline{1-4}
All Categories    & P      & R      & $F_{0.5}$  \\ \cline{1-4}
Norvig     & 0.5217 & 0.0355 & 0.1396 &  \\ \cline{1-4}
Enchant      & 0.2269 & 0.0411 & 0.1192 &  \\ \cline{1-4}
Jamspell & 0.4385 & 0.0449 & \textbf{0.1593} &  \\ \cline{1-4}
our      & 0.5116 & 0.0295 & 0.1198 &  \\ \cline{1-4}
\end{tabular}
}
\caption{Comparison of Grammatical Error Performance of Spellcheckers. Jamspell achieves the best score as previously suggested. \label{ta:sp_gec}}
\end{table}

\begin{table}[htb]
\resizebox{\columnwidth}{!}{%
\begin{tabular}{|l|l|l|l|l}
\cline{1-4}
R:SPELL    & P      & R & $F_{0.5}$ &  \\ \cline{1-4}
Norvig     & 0.5775 & 0.6357 & 0.5882 &  \\ \cline{1-4}
Enchant      & 0.316  & 0.6899 & 0.3544 &  \\ \cline{1-4}
Jamspell & 0.5336 & 0.6977 & 0.5599 &  \\ \cline{1-4}
our      & 0.6721 & 0.5297 & \textbf{0.6378} &  \\ \cline{1-4}
\end{tabular}
}
\caption{Comparison of spellcheckers on spelling. Our method outperforms other methods.\label{ta:sp_sp}}
\end{table}

\subsection{Nematus} \label{sec:nematus}
We trained Nematus using several different data sets. First, we trained using only the W\&I train set data,  we then added Lang8, FCE and Nucle data sources. Since Lang8 is significantly larger than W\&I train set, inspired by \citet{JunczysDowmunt2018ApproachingNG}, we upsampled W\&I 10 times  so that it will have more significant effect on the training process. This procedure improved results significantly (See Table \ref{ta:per_train}).

\begin{table}[htb]
\resizebox{\columnwidth}{!}{%
\begin{tabular}{|c|c|c|c|}
\hline
Training Data                                                                                                   & P      & R      & $F_{0.5}$   \\ \hline
    \begin{tabular}[c]{@{}c@{}}
    W\&I train set
    \end{tabular}
& 0.3187 & 0.1112 & 0.232  \\ \hline
    \begin{tabular}[c]{@{}c@{}}
    W\&I train set\\
    + lang8 + FCE
    \end{tabular}
& 0.4604 & 0.0742 & 0.225 \\ \hline
    \begin{tabular}[c]{@{}c@{}}
    W\&I train set\\
    (upsampled X 10)\\ + Lang8 + FCE + Nucle
    \end{tabular} & 0.4738 & 0.1529 & 0.333 \\ \hline
\end{tabular}
}
\caption{Nematus performance on W\&I dev set by training data. The use of more data improves the system, but only when the training from the domain is upsampled. \label{ta:per_train}}
\end{table}

\subsection{Synthetic Error Generation} \label{subsec:synth_exp}
\begin{table}[htb]
\resizebox{\columnwidth}{!}{%
\begin{tabular}{|c|c|c|cc}
\cline{1-3}
Data Source                                                           & \begin{tabular}[c]{@{}c@{}}Size (sentences)\end{tabular} & $F_{0.5}$   &  &  \\ \cline{1-3}
Gutenberg Books                                                       & 650,000                                                                   & 0.1483 &  &  \\ \cline{1-3}
Gutenberg Books                                                       & 7,000,000                                                                 & 0.1294 &  &  \\ \cline{1-3}
\begin{tabular}[c]{@{}c@{}}W\&I train set\end{tabular} & 1,300,000                                                                 & 0.1919 &  &  \\ \cline{1-3}
\end{tabular}
}
\caption{Size of synthetic datasets and Nematus scores when trained on them. \label{ta:synth}}
\end{table}
We also tried training Nematus over synthetic errors data. We generated errors using data from two different domains. Books from project Gutenberg and gold sentences from W\&I train set. Additionally, we varied data sizes and observed the effect on the results (See Table \ref{ta:synth}). These experiments show that relying on the source domain is crucial and it is best to generate data using text from similar domain. When using the synthetic W\&I train set we reached a score that is just a little lower than the score when training over W\&I train set directly (0.19 vs 0.23). This might suggest that there is potential in using synthetic data when combined with other data sets and promise for synthetic data methods for unsupervised GEC.

\subsection{Combining} \label{sec:merge}

The experiments regarding combining were performed on the dev set, which was not used for training the systems. The dev set was split to two randomly. The optimal selection of error-types and subsets to combine was done on one half, and we report system results on the second half. For example, when combining the output of the Nematus and Grammarly systems under 10 different fold partitions, the average $F_{0.5}$ improvement over the best of the two systems was 6.2 points, with standard deviation of 0.28 points.  

\paragraph{Improvement of a single tool.}

Even given a single system, we are able to improve the system's performance by eschewing predictions on low performing error types.  This filtering procedure has a minor effect and is exemplified in Table \ref{ta:filter}. While such findings are known to exist implicitly by the cycles of development \cite{choshen2018inherent}, and were suggested as beneficial for rule based and statistical machine translation systems when precision is 0 \cite{felice2014grammatical}, to the best of our knowledge we are the first to report those results directly, on non trivial precision with neural network based systems. In explicitly filtering corrections by error types we gain two additional benefits over the mere score improvement. First, the weak spots of the system are emphasized, and work might be directed to improving components or combining with a relevant strong system.  Second, the system itself is not discouraged or changed to stop producing those corrections. So, if future enhancement would improve this type of errors enough, it will show up in results, without discouraging smaller improvements done on the way.

\begin{table}[htb]
\resizebox{\columnwidth}{!}{%
\begin{tabular}{ |c||c|c|c|}
\hline
 System & P & R & $F_{0.5}$  \\
  \hline
  Language Tool  & 0.2905 &	0.1004	& 0.2107 \\
  Filtered Language Tool & 0.4005 &	0.0889  & 0.2355  \\
  \hline
  Grammarly & 	0.4846 &	0.1808	& 0.3627    \\
  Filtered Grammarly  & 0.5342	& 0.1715 &	0.3754   \\
  \hline 
  Nematus & 0.52 &	0.1751 &	0.373   \\
  Filtered Nematus & 0.554 & 	0.1647	& 0.3761  \\
 \hline
  
 \hline
\end{tabular}
}
\caption{Change in performance when avoiding hard errors. \label{ta:filter}}
\end{table}

\paragraph{Restricted track.} In Table \ref{ta:comb} we present the results of our shared task restricted track submission. The submission includes four Nematus models, our spellchecker, and Bert based system (\S \ref{subsec:bert}). This generated a 6 point improvement on the dev set of $f_{0.5}$ when compared the best standalone Nematus model.

\begin{table}[htb]
\resizebox{\columnwidth}{!}{%
\begin{tabular}{ |c||c|c|c|}
\hline
 System & P & R & $F_{0.5}$  \\
  \hline
  (1) Nematus1       &   0.4788	& 0.1544 & 0.3371   \\
  (2) Nematus2       &   0.4839	& 0.1583 & 0.3429    \\
  (3) Nematus3       &   0.4842	& 0.1489 & 	0.3338   \\
  (4) Nematus4       &   0.4843	& 0.1502 &	0.3352   \\
  (5) Spellchecker   &   0.5154	& 0.0308 &	0.1242   \\
  (6) Bert           &   0.0132	& 0.0147 &	0.0135   \\
  1+2                &   0.4972	& 0.1854 &	0.3721  \\
  1+2+3              &   \textbf{0.5095}	& 0.1904 & 	0.3816  \\
  1+2+3+4            &   0.4926	& 0.2017 & 	0.3824  \\
  1+2+3+4+5          &   0.5039	& 0.2233 &	0.4027  \\
  1+2+3+4+5+6        &   0.5029	& \textbf{0.2278} & \textbf{	0.4051}  \\
 \hline
\end{tabular}
}
\caption{Performance of systems and iterative combination of them. Combination improves both precision and recall even using low performing systems.\label{ta:comb}}
\end{table}
\paragraph{Off the shelf systems.} As can be seen in Table \ref{ta:off_shelf} when we combine the system with several off the self systems, we get 3 point improvement over the restricted baseline, and a 9 point improvement over the best standalone system. This implies there is a promise in combining existing approaches which we can't improve ourselves to harness some of their correction power. \footnote{Although some of the systems use only rules and non-parallel data, we did not include them in our submission to the restricted tracked, as we are not their originators.}

\begin{table}[htb]
\resizebox{\columnwidth}{!}{%
\begin{tabular}{ |c||c|c|c|}
\hline
 System & P & R & $F_{0.5}$  \\
  \hline
  (1) Restricted-best  &  0.5029	& 0.2278 & 	0.4051     \\
  (2) Language Tool    &  0.2699	& 0.0955 &	0.1977    \\
  (3) Grammerly        &  0.4783	& 0.1825 &	0.3612    \\
  (4) Jamspell         &  0.423	    & 0.0413 &  0.1484    \\
  1+2                  &  0\textbf{.5274}	& 0.2175 &	0.4105    \\
  1+2+3                &  0.522	    & \textbf{0.2656} &	\textbf{0.4375}    \\
  1+2+3+4              &  0.5221	& 0.2641 &	0.4367    \\
 \hline
\end{tabular}
}
\caption{Combining with off the shelf systems helps. \label{ta:off_shelf}}
\end{table}

\paragraph{Ensemble VS Combining models results.} \label{subsec:ensemble_compare}
Nematus has average ensembling built-in which enables inference over several RNN models by performing geometric average of the individual models' probability distributions.  Combining outperforms the built-in ensemble by almost 4 points (See Table \ref{ta:merge_vs_ensemble}). It is also important to note that while average ensemble improves precision, it reduces recall. Combination is balancing precision and recall, improving both, in a way that maximizes $F_{0.5}$. The last observation is far from trivial as most ways to combine systems would emphasize one or the other, e.g., piping would support mainly recall perhaps reducing precision. Lastly, combining is based on the types of errors and is linguistically motivated, and hence could be further improved by smart categorization and perhaps improvements of automatic detection \cite{Bryant2017AutomaticAA}.

\begin{table}[htb]
\resizebox{\columnwidth}{!}{%
\begin{tabular}{ |c||c|c|c|}
\hline
 System & P & R & $F_{0.5}$  \\
  \hline
  (1) Nematus RNN 1     & 0.4676 & 0.1157 & 0.2908   \\
  (2) Nematus RNN 2     & 0.4541 & 0.1223 & 0.2944    \\
  (3) Nematus RNN 3     & 0.484 & 0.1191 & 0.3002   \\
  (4) Nematus RNN 4     & 0.4839 & 0.1184 & 0.2991   \\
  1+2+3+4 ensemble      & \textbf{0.5577} & 0.1131 & 0.3122   \\
  1+2+3+4 combination  & 0.4861 & \textbf{0.166} & \textbf{0.3508}   \\
 \hline
\end{tabular}
}
\caption{Combining fares better compared to ensemble.}  \label{ta:merge_vs_ensemble}
\end{table}

\paragraph{Combining the shared task systems.}
After the completion of the competition test phase, several teams agreed to release their outputs on the dev and test set. We combined them using the entire dev set and submitted the results to the open phase of the restricted track for evaluation.  This achieves a 3.7 point improvement in $F_{0.5}$ and a 6.5 point improvement in precision over the best standalone results (See Table \ref{ta:bea_combined}). This means this combination is the best result currently known in the field as assessed by the BEA 2019 shared task.

\begin{table}[htb]
\resizebox{\columnwidth}{!}{%
\begin{tabular}{ |c||c|c|c|}
\hline
 System & P & R & $F_{0.5}$  \\
  \hline
  (1) UEDIN-MS     &  72.28 &	\textbf{60.12} & 	69.47  \\
  (2) Kakao\&Brain &  75.19 &	51.91 &	    69.00 \\  	  
  (3) Shuyao       &  70.17 & 	55.39 & 	66.61          \\
  (4) CAMB-CUED    &  66.75 &	53.93 & 	63.72 	       \\
  1+2              &  78.31	&   58.00 &	    \textbf{73.18} \\
  3+4              &  74.99	&   54.41 &     69.72          \\
  1+2+3+4          &  \textbf{78.74}	&   56.04 &  	72.84          \\
 \hline
\end{tabular}
}
\caption{Test set results when combining systems from the competition used as black boxes. The combination is the new state of the art.  \label{ta:bea_combined}}
\end{table}

%% file: discussion.tex
  \section{Conclusion and Future Work} \label{sec:discussion}
   
 In this paper, we have shown how combining multiple GEC systems, using a pure black-box approach, can improve state of the art results in the error correction task.  
 
 Additional variants of this combination approach can be further examined.  The approach can work with any disjoint partition  systems' corrections.  We can consider combining more than 2 systems at the same time, or we can consider more refined subsets of two systems.  For example, the set $H_{1 \setminus 2}$ of all the suggested corrections of $system1$ which were not suggested by $system2$, can be split to the two sets: $H_{1 overlapping 2}$ and  $H_{1 non-overlapping 2}$, the former containing corrections of system 1 which have an overlapping (but different) corrections by $system2$, and the later corrections of $system1$ which have no overlap with any annotation of $system2$.

 Several other approaches can be taken. The problem can be formulated as multiple-sequence to single sequence problem. The input sequences are the original text and $n$ system corrections.  The output sequence is the combined correction.  During training, the gold correction is used.   Given sufficient labeled data, it may be possible for such a system to learn subtle distinctions which may result in better combinations without relying on separating error types or iterative combinations.
 
 In addition, we harnessed Bert for GEC and showed a simple spellchecking mechanism yields competitive results to the leading spellcheckers.

%% file: GEC.bbl
\begin{thebibliography}{30}
\expandafter\ifx\csname natexlab\endcsname\relax\def\natexlab#1{#1}\fi

\bibitem[{Bryant and Briscoe(2018)}]{bryant-briscoe-2018-language}
Christopher Bryant and Ted Briscoe. 2018.
\newblock \href {https://doi.org/10.18653/v1/W18-0529} {Language model based
  grammatical error correction without annotated training data}.
\newblock In \emph{Proceedings of the Thirteenth Workshop on Innovative Use of
  {NLP} for Building Educational Applications}, pages 247--253, New Orleans,
  Louisiana. Association for Computational Linguistics.

\bibitem[{Bryant et~al.(2019)Bryant, Felice, Andersen, and Briscoe}]{bea2019}
Christopher Bryant, Mariano Felice, {\O}istein~E. Andersen, and Ted Briscoe.
  2019.
\newblock {The BEA-2019 Shared Task on Grammatical Error Correction}.
\newblock In \emph{Proceedings of the 14th Workshop on Innovative Use of NLP
  for Building Educational Applications}. Association for Computational
  Linguistics.

\bibitem[{Bryant et~al.(2017)Bryant, Felice, and
  Briscoe}]{Bryant2017AutomaticAA}
Christopher Bryant, Mariano Felice, and Ted Briscoe. 2017.
\newblock Automatic annotation and evaluation of error types for grammatical
  error correction.
\newblock In \emph{ACL}.

\bibitem[{Choshen and Abend(2018{\natexlab{a}})}]{choshen2018automatic}
Leshem Choshen and Omri Abend. 2018{\natexlab{a}}.
\newblock Automatic metric validation for grammatical error correction.
\newblock In \emph{Proceedings of the 56th Annual Meeting of the Association
  for Computational Linguistics (Volume 1: Long Papers)}, pages 1372--1382.

\bibitem[{Choshen and Abend(2018{\natexlab{b}})}]{choshen2018inherent}
Leshem Choshen and Omri Abend. 2018{\natexlab{b}}.
\newblock Inherent biases in reference-based evaluation for grammatical error
  correction and text simplification.
\newblock In \emph{ACL}.

\bibitem[{Choshen and Abend(2018{\natexlab{c}})}]{choshen2018reference}
Leshem Choshen and Omri Abend. 2018{\natexlab{c}}.
\newblock Reference-less measure of faithfulness for grammatical error
  correction.
\newblock \emph{arXiv preprint arXiv:1804.03824}.

\bibitem[{Dahlmeier and Ng(2012)}]{dahlmeier2012beam}
Daniel Dahlmeier and Hwee~Tou Ng. 2012.
\newblock A beam-search decoder for grammatical error correction.
\newblock In \emph{Proceedings of the 2012 Joint Conference on Empirical
  Methods in Natural Language Processing and Computational Natural Language
  Learning}, pages 568--578. Association for Computational Linguistics.

\bibitem[{Dahlmeier et~al.(2013)Dahlmeier, Ng, and Wu}]{dahlmeier2013building}
Daniel Dahlmeier, Hwee~Tou Ng, and Siew~Mei Wu. 2013.
\newblock Building a large annotated corpus of learner english: The nus corpus
  of learner english.
\newblock In \emph{Proceedings of the eighth workshop on innovative use of NLP
  for building educational applications}, pages 22--31.

\bibitem[{Dale and Kilgarriff(2011)}]{dale2011helping}
Robert Dale and Adam Kilgarriff. 2011.
\newblock Helping our own: The hoo 2011 pilot shared task.
\newblock In \emph{Proceedings of the 13th European Workshop on Natural
  Language Generation}, pages 242--249. Association for Computational
  Linguistics.

\bibitem[{Devlin et~al.(2018)Devlin, Chang, Lee, and
  Toutanova}]{devlin2018bert}
Jacob Devlin, Ming-Wei Chang, Kenton Lee, and Kristina Toutanova. 2018.
\newblock Bert: Pre-training of deep bidirectional transformers for language
  understanding.
\newblock \emph{arXiv preprint arXiv:1810.04805}.

\bibitem[{Dlugolinsk{\`y} et~al.(2013)Dlugolinsk{\`y}, Krammer, Ciglan,
  Laclav{\'\i}k, and Hluch{\`y}}]{dlugolinsky2013combining}
{\v{S}}tefan Dlugolinsk{\`y}, Peter Krammer, Marek Ciglan, Michal
  Laclav{\'\i}k, and Ladislav Hluch{\`y}. 2013.
\newblock Combining named enitity recognition tools.
\newblock \emph{Making Sense of Microposts (\# MSM2013)}.

\bibitem[{Felice and Yuan(2014)}]{felice2014generating}
Mariano Felice and Zheng Yuan. 2014.
\newblock Generating artificial errors for grammatical error correction.
\newblock In \emph{Proceedings of the Student Research Workshop at the 14th
  Conference of the European Chapter of the Association for Computational
  Linguistics}, pages 116--126.

\bibitem[{Felice et~al.(2014)Felice, Yuan, Andersen, Yannakoudakis, and
  Kochmar}]{felice2014grammatical}
Mariano Felice, Zheng Yuan, {\O}istein~E Andersen, Helen Yannakoudakis, and
  Ekaterina Kochmar. 2014.
\newblock Grammatical error correction using hybrid systems and type filtering.
\newblock In \emph{Proceedings of the Eighteenth Conference on Computational
  Natural Language Learning: Shared Task}, pages 15--24.

\bibitem[{Goldberg(2019)}]{DBLP:journals/corr/abs-1901-05287}
Yoav Goldberg. 2019.
\newblock \href {http://arxiv.org/abs/1901.05287} {Assessing bert's syntactic
  abilities}.
\newblock \emph{CoRR}, abs/1901.05287.

\bibitem[{Granger(1998)}]{Granger1998TheCL}
Sylviane Granger. 1998.
\newblock The computerized learner corpus: a versatile new source of data for
  sla research.

\bibitem[{Grundkiewicz and Junczys-Dowmunt(2018)}]{grundkiewicz2018near}
Roman Grundkiewicz and Marcin Junczys-Dowmunt. 2018.
\newblock Near human-level performance in grammatical error correction with
  hybrid machine translation.
\newblock \emph{arXiv preprint arXiv:1804.05945}.

\bibitem[{Jiang et~al.(2016)Jiang, Banchs, and Li}]{jiang2016evaluating}
Ridong Jiang, Rafael~E Banchs, and Haizhou Li. 2016.
\newblock Evaluating and combining name entity recognition systems.
\newblock In \emph{Proceedings of the Sixth Named Entity Workshop}, pages
  21--27.

\bibitem[{Junczys-Dowmunt et~al.(2018)Junczys-Dowmunt, Grundkiewicz, Guha, and
  Heafield}]{JunczysDowmunt2018ApproachingNG}
Marcin Junczys-Dowmunt, Roman Grundkiewicz, Shubha Guha, and Kenneth Heafield.
  2018.
\newblock Approaching neural grammatical error correction as a low-resource
  machine translation task.
\newblock In \emph{NAACL-HLT}.

\bibitem[{Koehn et~al.(2007)Koehn, Hoang, Birch, Callison-Burch, Federico,
  Bertoldi, Cowan, Shen, Moran, Zens et~al.}]{koehn2007moses}
Philipp Koehn, Hieu Hoang, Alexandra Birch, Chris Callison-Burch, Marcello
  Federico, Nicola Bertoldi, Brooke Cowan, Wade Shen, Christine Moran, Richard
  Zens, et~al. 2007.
\newblock Moses: Open source toolkit for statistical machine translation.
\newblock In \emph{Proceedings of the 45th annual meeting of the association
  for computational linguistics companion volume proceedings of the demo and
  poster sessions}, pages 177--180.

\bibitem[{Levenshtein(1966)}]{Levenshtein1966BinaryCC}
Vladimir~I Levenshtein. 1966.
\newblock Binary codes capable of correcting deletions, insertions, and
  reversals.
\newblock In \emph{Soviet physics doklady}, volume~10, pages 707--710.

\bibitem[{Mizumoto et~al.(2011)Mizumoto, Komachi, Nagata, and
  Matsumoto}]{mizumoto2011mining}
Tomoya Mizumoto, Mamoru Komachi, Masaaki Nagata, and Yuji Matsumoto. 2011.
\newblock Mining revision log of language learning sns for automated japanese
  error correction of second language learners.
\newblock In \emph{Proceedings of 5th International Joint Conference on Natural
  Language Processing}, pages 147--155.

\bibitem[{Napoles et~al.(2017)Napoles, Sakaguchi, and
  Tetreault}]{napoles2017jfleg}
Courtney Napoles, Keisuke Sakaguchi, and Joel~R. Tetreault. 2017.
\newblock Jfleg: A fluency corpus and benchmark for grammatical error
  correction.
\newblock In \emph{EACL}.

\bibitem[{Rozovskaya and Roth(2011)}]{rozovskaya2011Algorithm}
Alla Rozovskaya and Dan Roth. 2011.
\newblock Algorithm selection and model adaptation for esl correction tasks.
\newblock In \emph{ACL}.

\bibitem[{Rozovskaya and Roth(2016)}]{rozovskaya2016grammatical}
Alla Rozovskaya and Dan Roth. 2016.
\newblock Grammatical error correction: Machine translation and classifiers.
\newblock In \emph{Proceedings of the 54th Annual Meeting of the Association
  for Computational Linguistics (Volume 1: Long Papers)}, volume~1, pages
  2205--2215.

\bibitem[{Rozovskaya et~al.(2014)Rozovskaya, Roth, and
  Srikumar}]{rozovskaya2014correcting}
Alla Rozovskaya, Dan Roth, and Vivek Srikumar. 2014.
\newblock Correcting grammatical verb errors.
\newblock In \emph{Proceedings of the 14th Conference of the European Chapter
  of the Association for Computational Linguistics}, pages 358--367.

\bibitem[{Ruiz and Poibeau(2015)}]{ruiz2015combining}
Pablo Ruiz and Thierry Poibeau. 2015.
\newblock Combining open source annotators for entity linking through weighted
  voting.
\newblock In \emph{Joint Conference on Lexical and Computational Semantics (*
  SEM 2015)}, pages 211--215.

\bibitem[{Sennrich et~al.(2017)Sennrich, Firat, Cho, Birch, Haddow, Hitschler,
  Junczys-Dowmunt, L\"{a}ubli, Miceli~Barone, Mokry, and
  Nadejde}]{sennrich2017EACLDemo}
Rico Sennrich, Orhan Firat, Kyunghyun Cho, Alexandra Birch, Barry Haddow,
  Julian Hitschler, Marcin Junczys-Dowmunt, Samuel L\"{a}ubli, Antonio~Valerio
  Miceli~Barone, Jozef Mokry, and Maria Nadejde. 2017.
\newblock \href {http://aclweb.org/anthology/E17-3017} {Nematus: a toolkit for
  neural machine translation}.
\newblock In \emph{Proceedings of the Software Demonstrations of the 15th
  Conference of the European Chapter of the Association for Computational
  Linguistics}, pages 65--68, Valencia, Spain. Association for Computational
  Linguistics.

\bibitem[{Sennrich et~al.(2015)Sennrich, Haddow, and
  Birch}]{sennrich2015neural}
Rico Sennrich, Barry Haddow, and Alexandra Birch. 2015.
\newblock Neural machine translation of rare words with subword units.
\newblock \emph{arXiv preprint arXiv:1508.07909}.

\bibitem[{Vaswani et~al.(2017)Vaswani, Shazeer, Parmar, Uszkoreit, Jones,
  Gomez, Kaiser, and Polosukhin}]{Vaswani2017Attention}
Ashish Vaswani, Noam Shazeer, Niki Parmar, Jakob Uszkoreit, Llion Jones,
  Aidan~N. Gomez, Lukasz Kaiser, and Illia Polosukhin. 2017.
\newblock Attention is all you need.
\newblock In \emph{NIPS}.

\bibitem[{Zhao et~al.(2019)Zhao, Wang, Shen, Jia, and Liu}]{zhao2019improving}
Wei Zhao, Liang Wang, Kewei Shen, Ruoyu Jia, and Jingming Liu. 2019.
\newblock Improving grammatical error correction via pre-training a
  copy-augmented architecture with unlabeled data.
\newblock \emph{arXiv preprint arXiv:1903.00138}.

\end{thebibliography}
